\documentclass[
]{ceurart}

\usepackage{listings}
\begin{document}

\copyrightyear{2025}
\copyrightclause{Copyright for this paper by its authors.
  Use permitted under Creative Commons License Attribution 4.0
  International (CC BY 4.0).}

\conference{AIDA2J at JURIX25: Worskop on AI for Access to Justice, Dispute Resolution, and Data Access,
  December 09, 2025, Torino, Italy}


\title{Glitter: Visualizing Lexical Surprisal for Readability in Administrative Texts}
\author{Jan Černý}[orcid=0009-0006-4873-8233, email=jc@ucw.cz]
\fnmark[1]
\author{Ivana Kvapilíková}[orcid=0000-0003-1479-3294, email=kvapilikova@ufal.mff.cuni.cz]
\author{Silvie Cinková}[orcid=0000-0001-9455-9013, email=conkova@ufal.mff.cuni.cz]

\address{Charles University, Faculty of Mathematics and Physics}

\cortext[1]{Corresponding author.}


\begin{abstract}
This work investigates how measuring information entropy of text can be used to estimate its readability. We propose a visualization framework that can be used to approximate information entropy of text using multiple language models and visualize the result. The end goal is to use this method to estimate and improve readability and clarity of administrative or bureaucratic texts. Our toolset is available as a libre software on \url{https://github.com/ufal/Glitter}.
\end{abstract}

\maketitle

\section{Introduction}

The readability of legal texts has long been debated in legal scholarship and is regarded as a central aspect of access to justice. This concern has been repeatedly emphasized within the Plain Language movement \cite{kimble2023writing}.

This paper proposes the development of a natural language processing (NLP) tool designed to assess the comprehensibility of texts and to assist authors in producing writing that is more accessible to readers. In particular, we focus on written communication between citizens and public authorities. The motivation for this work arises from the observation that administrative or bureaucratic texts with legal terminology are frequently difficult to understand, particularly for individuals who interact with public authorities but lack legal education. Enhancing the readability of such texts could therefore contribute to more effective communication and greater accessibility of information that is important, necessary, and often legally binding.

 It was shown that readability is hindered by words that are difficult to anticipate from existing context and thus impose a greater cognitive load on the reader \cite{LEVY20081126}. In this study, we use language models to measure text predictability and we argue that also passages that are too predictable for language models can be difficult to read. Our visualization tool called Glitter is publicly available (\url{https://quest.ms.mff.cuni.cz/ponk-app2/}) and can be used to visualize how information is spread across the input text. In the context of writing an administrative document, it can serve as a guide on what passages are likely difficult to read. %The contributions of our work are the following: we show that, contrary to previous research,  % Our focus is on administrative documents intended for readers who often lack specialized legal or bureaucratic knowledge.

\section{Related Work}
We build upon prior work on the impact of lexical surprisal on text readability \cite{DEMBERG2008193, pimemtel2023, wilcox-etal-2023-testing, SunWang2025ComputationalSentenceLevel} where probabilistic language models are used to quantify how expected a word is in its context. These works are based on the surprisal theory \cite{hale-2001-probabilistic, LEVY20081126} which links text predictability with cognitive processing difficulty in reading. In contrast to previous work, we point out that there are certain kinds of texts that are highly predictable for the language models but incomprehensible for human readers.

Another line of research has used probabilities from language models to detect automatically generated text. \cite{gltr} showed that automatically generated texts exhibit overall low surprisal as a result of the fact that text is generated by sampling from a probability distribution. In contrast, natural texts are less likely when analyzed through the lens of a language model. We re-implement and extend their text visualization toolkit and use it to visualize surprisal in administrative documents.

\subsection{Language modeling}
Large language models (LLMs) are used for a wide variety of tasks, often through prompting strategies that condition the model to generate task-specific outputs. Although they can be adapted to many applications, their underlying mechanism remains next-token prediction: modeling the probability distribution of upcoming tokens given preceding context. In this work, we return to this core modeling objective and use it as a framework for analyzing texts themselves. By examining how well an LLM can model a given text, we aim to gain insight into the text’s properties and its overall predictability and readability.

Large language models assign high probability to text sequences that strongly resemble patterns present in their training data. When a model identifies a passage as extremely likely (so likely that it appears to recall it nearly verbatim) this often indicates that the passage corresponds to a fixed expression, a formulaic phrase or a frequently repeated citation. In the context of administrative and legal documents, such highly predictable spans typically reflect fixed legal terminology, standardized procedural clauses or \textit{boilerplate} language.

Texts that rely heavily on such formulaic segments may be more difficult for non-expert readers to understand. These expressions carry specialized legal meanings, often compressed into rigid, citation-like phrasing that does not align with everyday language use. As a result, a high density of these highly model-predictable passages can serve as a proxy for the presence of technical legal language, which is known to impede readability and comprehension among lay readers.

Our visualization tool extracts probabilities from a language model and detects such passages as sequences of words where each following word is highly likely.

\section{Methodology}

\subsection{Lexical surprisal}
Lexical surprisal refers to the degree of unexpectedness or unpredictability of a word within a given context, sentence, or text. It is a measure of how surprising or unusual a word is based on linguistic norms, probabilities, or reader's expectations given the preceding text. %In this work, we make the assumption that the level of lexical surprisal in text could be used to estimate text comprehensibility and that it can be measured by a probabilistic language model trained on vast amounts of text.
In NLP terms, lexical surprisal corresponds to the inverse conditional probability of each word given its left context. %The lower the probability, the higher the surprisal experienced by the reader.

 The linguistic concept of lexical surprisal is closely linked to the concept of information entropy as formulated in information theory by Claude Shannon \cite{shannon}. Surprisal refers to the unexpectedness of a word: the less probable a word is, the higher its surprisal value. Entropy, in turn, is defined as the average uncertainty of a probability distribution which can be expressed mathematically as

 $$H(X) = -\sum_{i=1}^{n} p_i \log \left(p_i\right)$$

%In linguistic terms, entropy captures the overall unpredictability of a text. This uneven distribution makes the text more difficult to process and comprehend, since the reader must constantly adjust to varying levels of predictability. By contrast, a text with low entropy has a more uniform and balanced distribution of surprisal, meaning that word occurrences are relatively predictable. Such texts are easier to follow, though they may also appear monotonous or less informative, since low entropy often corresponds to redundancy.
A text with low entropy has a lower overall level of surprisal, meaning that word occurrences are relatively predictable. Such texts are easier to follow, though they may also appear monotonous or less informative, since low entropy often corresponds to redundancy.

\subsection{Implementation}
NLP has numerous options how to estimate text probability, ranging from n-gram language models, through small neural networks trained on moderately sized text data collections to large language models trained on virtually all text data on the internet. %In this work, we use two types of pre-trained language models specialized in the Czech language data -- a causal Czech-GPT-2-XL\footnote{\url{https://huggingface.co/BUT-FIT/Czech-GPT-2-XL-133k}} model and an unmasking RobeCzech\footnote{\url{https://huggingface.co/ufal/robeczech-base}} model.
In this work, we use the GPT-2\footnote{\url{https://huggingface.co/openai-community/gpt2}}, a 1.5B parameter Transformer model trained on a collection of millions of webpages comprising $\sim$40 GB of text data \cite{Radford2019LanguageMA}. We also experimented with unmasking models \cite{devlin-etal-2019-bert, robeczech}, but concluded that such models are not suitable for our scenario, as their bidirectional training objective does not align with the sequential manner in which humans process text.  Commercial providers of modern LLMs such as OpenAI and Google do not expose token-level probabilities through their APIs, which prevents direct access to the underlying logits. For this reason, our experiments rely on open-source models, where the full probability distribution can be retrieved and manipulated.

%, trained on 
%We model lexical surprisal as informational entropy.

%$$H(X) = -\sum_{i=1}^{n} p_i \log \left(p_i\right)$$

%where $p_i$ represents the probability of the $i$-th word in the context of text. We model this probability using transformer models, in our case GPT and BERT models.

\begin{figure}[]\centering
	\includegraphics[width=\textwidth]{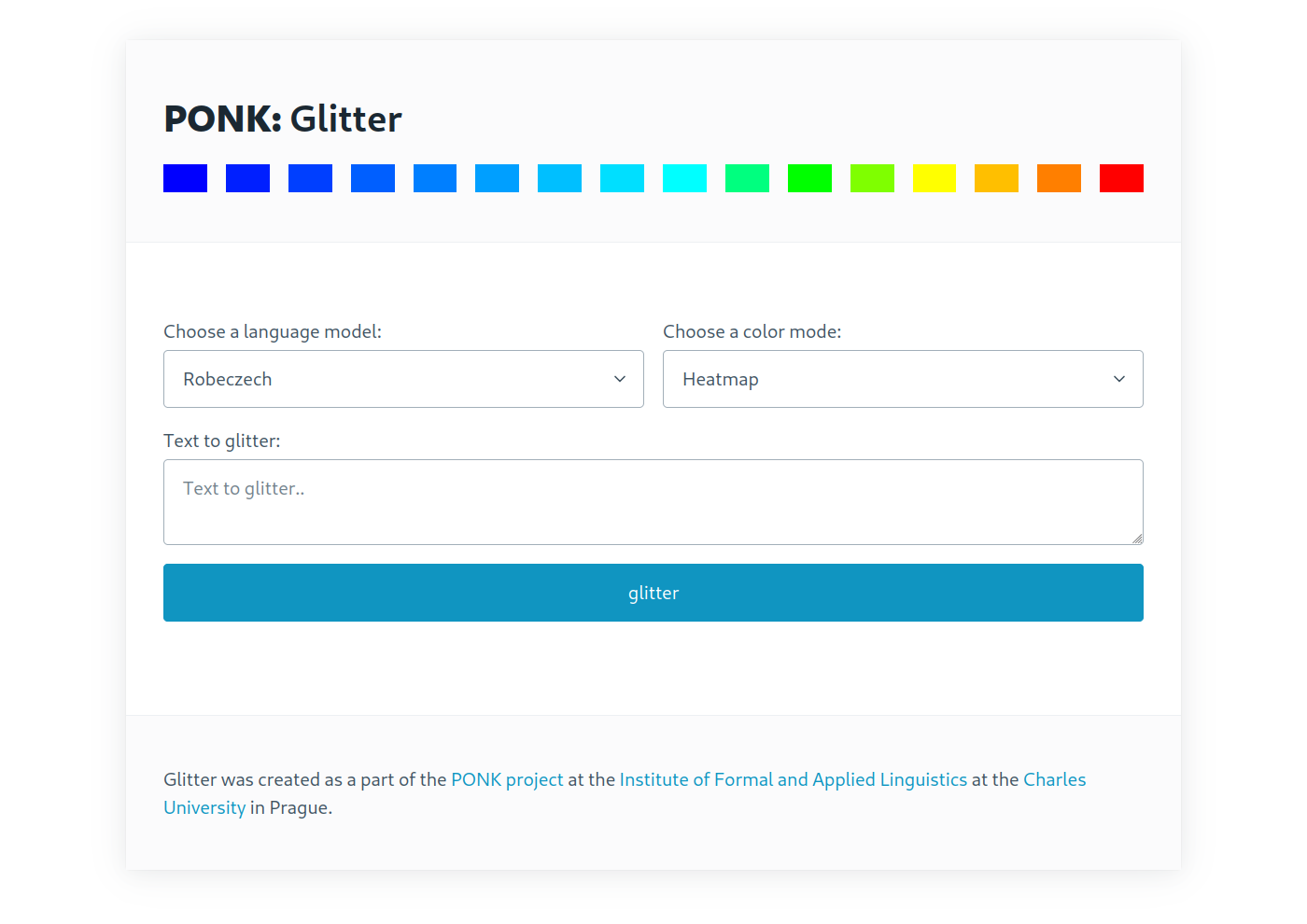}
    \vspace{-2em}
	\caption{\texttt{Glitter} web user interface in light mode}
	\label{fig:glitter-webui-light}
\end{figure}

%\subsection{Implementation}
%This section describes the pipeline of text through functions which each make specific step of glittering process. Details of implementation of these functions can differ per model because it depends on properties of LLM they are using, but all of them do the same in principle.
The following pipeline describes the steps we take in the probability estimation and surprisal annotation:

\begin{enumerate}
    \item \textbf{Input Conversion}\\
     We normalize the text, tokenize it and split into subword tokens to match the vocabulary of the pre-trained model we use.

    \item \textbf{Running through language model}\\
     We pass the input through a pre-trained language model and store the output logits.
     We normalize the logits through a softmax function to obtain a token probability distribution for each position of the text.

    \item \textbf{Rank and Probability of Original Token}\\
     We extract the probability of each token of the text and calculate surprisal as defined in Equation 1. We also extract the top 5 token candidates for each position of the text. For words that are segmented into subword tokens, the overall probability is computed as the product of the conditional probabilities of the constituent subword tokens, following the chain rule of probability.

    \item \textbf{Result Construction}\\
     We visualize the output by assigning colors to each token according to its surprisal. The ranking is divided into 16 buckets, with bucket sizes increasing as the predictability increases. This design choice allows us to highlight fine-grained differences among the most probable tokens, while paying less attention to the long tail of the distribution. The color scale is inspired by a thermal camera palette with \emph{warmer} colors representing higher surprisal values. We also show the user which top 5 words are expected at each given position.
\end{enumerate}

%\subsection{Tool}
%We adopted terminology from CITACE NA PUVODNI GLITTER. Glittering refers to process of calculating lexical surprisal.
%Glittered text is text where every word or token has visualized information about its lexical surprisal.
%Glitter is software tool that can be used to glitter text.

We call our visualization tool \texttt{Glitter}, referencing the \texttt{GLTR} framework we build upon \cite{gltr}. The annotation of text with token probabilities and top candidates can thus be called \textit{``glittering"}. The frontend of our application is illustrated in Figure \ref{fig:glitter-webui-light}. %in context of this thesis refers to process of creating glittered text from input text. ``Glittered text" is tokenized text extended about lexical surprisal. ``Glitter" is tool to glitter text. 
Our implementation  supports multiple language model architectures and has no limit of length of text that can be processed. It also provides graphical interface and command line user interface. The source code is available on \url{github.com/ufal/Glitter}. The tool was created as part of the PONK project\footnote{\url{https://ufal.mff.cuni.cz/ponk}}
 and is integrated in a writing-assistance application for the Czech language accessible at \url{https://quest.ms.mff.cuni.cz/ponk/}.

\section{Demonstration}
 
%We now present the visualized results for two \textit{glittered} texts, purposefully selecting an extreme case to provide clearer contrast. Lexical surprisal was computed for each token in the model, and the visualization employs a color scheme inspired by thermal imaging: tokens appear “warmer” as their lexical surprisal increases.

%In the example shown in Figure \ref{fig:legal1}, the text is particularly difficult to read, and the language model achieves a low accuracy in token prediction. This is reflected in long stretches of words with high lexical surprisal.

\begin{figure}[]\centering
	\includegraphics[width=\textwidth]{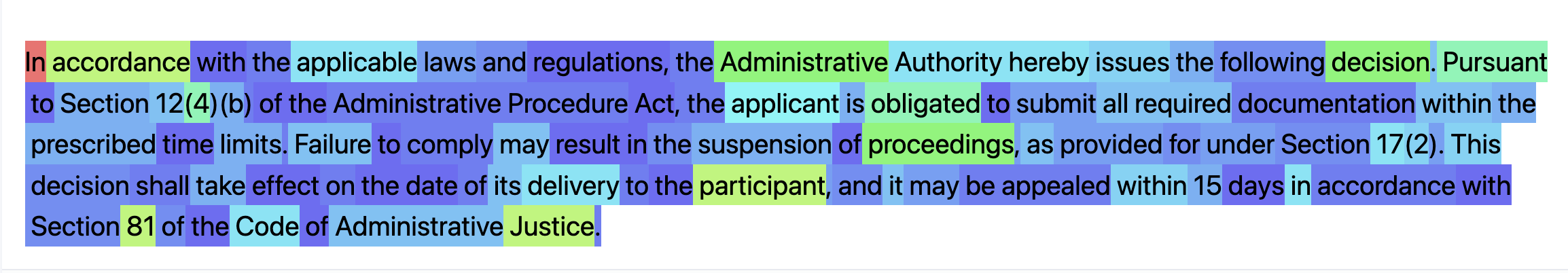}
	\caption{\textit{Glittered} administrative text}
	\label{fig:legal2}
\end{figure}

%We now  present the visual outputs of the \textit{glittering} process. As you can see in example in Figure \ref{fig:legal2}, the tokens with high lexical surprisal are distributed evenly throughout the passage. In this case, words with a higher information value tend to exhibit greater lexical surprisal. On the other hand, tokens are uninformative or serve primarily syntactic functions and have a much smaller lexical surprisal. The word "counterweights" has the highest lexical surprisal in this text. language model ranked it as 266303 possible words. The most probable ware extimated these words in this order: the,a,we,I,you. To show estimation of some word with low lexical surprisal, word "Information" on third line was ranked as first by language model, and other most probable options were: was, could, would, might.

We now present the outputs of our visualization tool applied to a text that is inherently difficult to read (\ref{fig:legal2}). The most predictable passages are colored in blue whereas the words with more information value are highlighted in green. Red color is reserved for words with the highest surprisal. In this example, it is only the first word which is impossible to predict due to missing preceding context.

   In general, words with greater informational value tend to demonstrate higher lexical surprisal and are essential for the text to convey meaning. Conversely, tokens that are less informative or serve primarily syntactic functions exhibit considerably lower surprisal values. However, we also observed that formulaic expressions such as \textit{"submit all required documentation within the prescribed time limits"} or \textit{"decision shall take effect on the date of its delivery to the participant"} exhibit low surprisal as well, although they could be conveyed in a more comprehensible manner. This outcome reflects the nature of the underlying model, which has been extensively trained on legal texts. However, it is in contract with previous  work which uses a language model as a proxy for a human reader.

   It shows that one limitation of measuring lexical surprisal with large pretrained language models is that these models may be too powerful for the intended purpose. Since models such as GPT-2 are trained on vast amounts of internet data, they have effectively been exposed to enormous quantities of technical, bureaucratic, and legal documents. As a result, their predictions often reflect specialized knowledge that far exceeds the linguistic competence of the target readership. In other words, the surprisal values produced by such models may underestimate the actual cognitive difficulty faced by ordinary readers, particularly when encountering complex or domain-specific terminology.
   
   %This is mostly because the model could not tune to the legal domain since it is the beginning of text. Later in the passage we observe that technical terms that 
%By contrast, the most probable tokens estimated by the model for that position were \textit{``the''}, \textit{``a''}, \textit{``we''}, \textit{``I''}, and \textit{``you''}. To illustrate an example of a word with low lexical surprisal, the token \textit{``information''} on the third line was ranked as the most probable candidate, followed closely by alternatives such as \textit{``was''}, \textit{``could''}, \textit{``would''}, and \textit{``might''}.

We performed a limited manual evaluation of the Glitter tool on the KUKY dataset\footnote{\url{https://ufal.mff.cuni.cz/grants/ponk/kuky}}
 of administrative texts, and our qualitative analysis showed distinct local patterns of lexical surprisal in versions post-edited in a reader-oriented manner, with the aim of improving clarity and comprehension. We foresee a more systematic evaluation in the future work.

%An alternative approach would be to employ simpler statistical models, such as n-gram language models, which rely only on surface-level co-occurrence statistics within limited contexts. While these models lack the world knowledge and deep contextual understanding of modern neural models, they may provide surprisal estimates that are more interpretable and closer to the actual processing demands experienced by non-expert readers. In this sense, the relative simplicity of n-gram models could serve as an advantage when the goal is to study text readability rather than to maximize predictive accuracy.
\section{Conclusion}

In conclusion, our results suggest that measuring lexical surprisal has strong potential as a metric for estimating information distribution in text and assessing readability. We believe that this approach warrants further attention and systematic experimentation.

There are several directions for future work:

\begin{itemize}
\item Developing multi-model estimation approaches with the focus on awider range of model architectures, including n-gram models.
%\item Extending support for and experimenting with a wider range of model architectures.
\item Incorporating system prompts to allow users to adapt model behavior, for example to reflect a specific vocabulary or target audience.
\item Applying the \texttt{Glitter} tool to large corpora of legal or bureaucratic texts aimed at non-expert readers and conducting a systematic evaluation of their readability.
\end{itemize}

\vspace{1 em}\noindent \textbf{Acknowledgements}
The authors gratefully acknowledge support from the Technology Agency of the Czech Republic, grant TQ01000526. The research reported in the present contribution has been using language resources developed, stored and distributed by the LINDAT/CLARIAH-CZ project of the Ministry of Education, Youth and Sports of the Czech Republic (LM2023062).
\bibliography{sample-1col}

\end{document}